\documentclass[letterpaper]{article} 
\usepackage{aaai25}  
\usepackage{times}  
\usepackage{helvet}  
\usepackage{courier}  
\usepackage[hyphens]{url}  
\usepackage{graphicx} 
\usepackage{natbib}  
\usepackage{caption} 
\usepackage{algorithm}
\usepackage[colorlinks,
linkcolor=blue,
citecolor=blue
]{hyperref}
\usepackage{color}

\usepackage{algorithmic}
\usepackage{multirow}
\usepackage{booktabs}
\usepackage{amsthm,amsmath,amssymb}
\usepackage{mathrsfs}
\usepackage{newfloat}
\usepackage{listings}
\DeclareCaptionStyle{ruled}{labelfont=normalfont,labelsep=colon,strut=off} 
\floatstyle{ruled}
\newfloat{listing}{tb}{lst}{}
\floatname{listing}{Listing}
\title{ISPDiffuser: Learning RAW-to-sRGB Mappings with Texture-Aware Diffusion Models and Histogram-Guided Color Consistency}
\author{
    Yang Ren\textsuperscript{\rm 1,\rm 4,\equalcontrib},
    Hai Jiang\textsuperscript{\rm 1,\rm 4,\equalcontrib},
    Menglong Yang\textsuperscript{\rm 1,\rm 2,\footnotemark[2]},
    Wei Li\textsuperscript{\rm 1,\rm 2},
    Shuaicheng Liu\textsuperscript{\rm 3,\rm 4,\footnote{Corresponding authors.}}
}
\affiliations{
    \textsuperscript{\rm 1}School of Aeronautics and Astronautics, Sichuan University\\
    \textsuperscript{\rm 2}Key Laboratory of Advanced Spatial Mechanism and Intelligent Spacecraft, Sichuan University \\
    \textsuperscript{\rm 3}University of Electronic Science and Technology of China \\
    \textsuperscript{\rm 4}Megvii Technology\\

    \{renyang@stu.,jianghai@stu.,mlyang@,li.wei@\}scu.edu.cn, liushuaicheng@uestc.edu.cn}

\usepackage{bibentry}

\begin{document}

\maketitle

\begin{abstract}
RAW-to-sRGB mapping, or the simulation of the traditional camera image signal processor (ISP), aims to generate DSLR-quality sRGB images from raw data captured by smartphone sensors. Despite achieving comparable results to sophisticated handcrafted camera ISP solutions, existing learning-based methods still struggle with detail disparity and color distortion. In this paper, we present ISPDiffuser, a diffusion-based decoupled framework that separates the RAW-to-sRGB mapping into detail reconstruction in grayscale space and color consistency mapping from grayscale to sRGB. Specifically, we propose a texture-aware diffusion model that leverages the generative ability of diffusion models to focus on local detail recovery, in which a texture enrichment loss is further proposed to prompt the diffusion model to generate more intricate texture details. Subsequently, we introduce a histogram-guided color consistency module that utilizes color histogram as guidance to learn precise color information for grayscale to sRGB color consistency mapping, with a color consistency loss designed to constrain the learned color information. Extensive experimental results show that the proposed ISPDiffuser outperforms state-of-the-art competitors both quantitatively and visually.
\end{abstract}

%
\begin{links}
\link{Code}{https://github.com/RenYangSCU/ISPDiffuser}
\end{links}
\section{Introduction}\label{sec: intro}
The image signal processor (ISP) pipeline is a fundamental process for converting RAW data captured by camera sensors into sRGB images that are perceivable by the human eye. Traditionally, the ISP pipeline in cameras consists of a series of discrete processes, including demosaicing, denoising, white balance, gamma correction, and color correction, each of which necessitates complex and extensive manual parameter tuning~\cite{ramanath2005color}. 
\begin{figure}[!t]
    \centering
    \includegraphics[width=\linewidth]{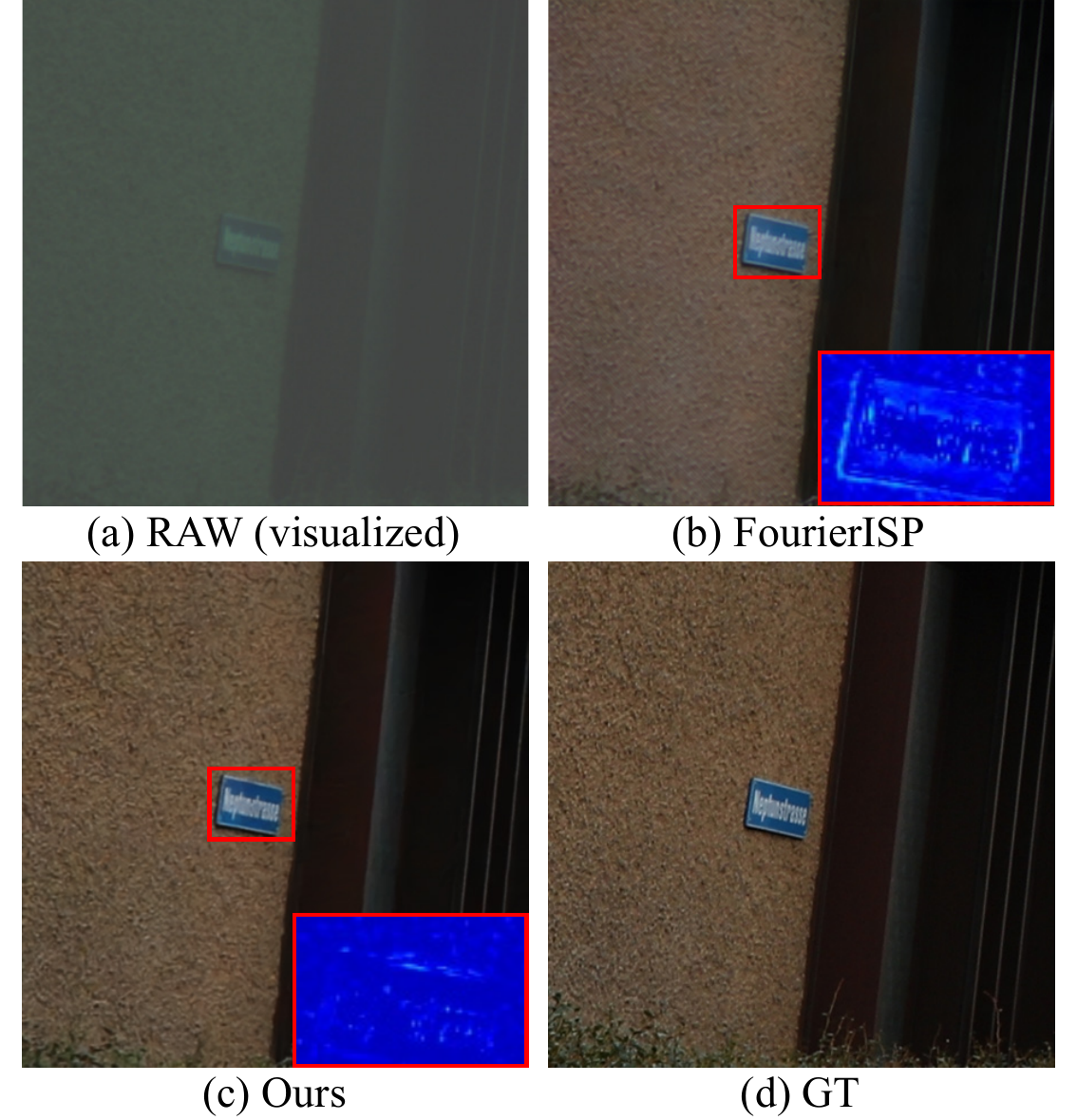}
    \caption{Visual comparison with the previous state-of-the-art method FourierISP~\cite{FourierISP}. Our approach exhibits better local detail reconstruction (the red boxes show the content difference between generated images and GT images) and global color consistency mapping capabilities.}
    \label{fig: teaser}
\end{figure}

With the rapid advancement in mobile photography, smartphones have become the dominant photography tool due to their convenience and portability. However, due to limitations in aperture and sensor size, mobile devices generally produce images with lower quality than those captured by DSLR cameras. To address this disparity, there has been growing interest in developing deep learning-based ISP models that can convert RAW data captured by mobile sensors into sRGB images with DSLR-like quality~\cite{ignatov2021learned}. Existing deep ISP solutions focus on either compensating for the misalignment caused by different capture equipment of the training pairs~\cite{LiteISPNet} or treating the RAW-to-sRGB mapping solely as a color mapping task~\cite{FourierISP, ignatov2020replacing, AWNet}. Despite the remarkable progress, a notable limitation persists: most of these models are based on convolutional neural networks (CNNs), which are naturally limited by their inherent locality restriction, leading to local detail disparity and global color distortion. As shown in Fig.~\ref{fig: teaser}(b) and (d), the previous state-of-the-art method FourierISP~\cite{FourierISP} produces blurred details and color distortion to degrade visual quality. 

Recently, generative model-based methods~\cite{GAN_vision, diffusion_vision} have emerged as promising approaches for various low-level vision tasks to achieve better perceptual quality, where diffusion models~\cite{DDPM, DDIM} have gained significant attention due to their impressive generative capabilities and advantages over previous generative models, such as generative adversarial networks (GANs) and variational autoencoders (VAEs), by avoiding instability and mode-collapse problems. DiffRAW~\cite{diffraw} proposed a diffusion-based framework for RAW-to-sRGB mappings that utilizes the sRGB images produced by LiteISPNet~\cite{LiteISPNet} as the color-position preserving condition, which however would constrain the learned color and detail distributions similar to LiteISPNet. Moreover, diffusion models, despite their superior high-frequency detail information generation capability, are usually biased in low-frequency information generation such as color and illumination~\cite{diffsuion_bias}, which presents challenges in handling local detail reconstruction and global color mapping simultaneously for RAW-to-sRGB mappings. 

To this end, we proposed a diffusion-based decoupled framework, named ISPDiffuser, which separates RAW-to-sRGB mappings into detail reconstruction in grayscale space and color consistency mapping from grayscale to sRGB to achieve visually satisfactory results. Specifically, we propose a texture-aware diffusion model (TADM) that leverages the generative ability of diffusion models to focus on details reconstruction without concern for color information, in which the encoded feature of the grayscale version of the corresponding sRGB image is taken as input to perform diffusion processes with the guidance of encoded RAW feature and newly proposed texture enrichment loss. Subsequently, we present a histogram-guided color consistency module (HCCM) that employs the color histogram~\cite{color_histogram} as guidance to learn precise color information to transform the gray feature generated by TADM into the sRGB feature with consistent colors as DSLR images, with a color consistency loss formulated to supervise the learned color information. As shown in Fig.~\ref{fig: teaser}(c), our method is capable of generating sRGB images with more distinct details and vivid color, being more visually pleasant. Extensive experiments demonstrate that our method outperforms existing state-of-the-art competitors quantitatively and visually. 

Our contributions can be summarized as follows:
\begin{itemize}
    \item We propose a diffusion-based decoupled framework, dubbed ISPDiffuser, which performs detail reconstruction and color consistency mapping separately to achieve visually satisfactory RAW-to-sRGB mappings. 
    \item We propose a texture-aware diffusion model that leverages the generative ability of diffusion models to focus on detail reconstruction, as well as a histogram-guided color consistency module that utilizes the color histogram to learn stable global color mapping. 
    \item Extensive experiments demonstrate that our method outperforms existing state-of-the-art competitors and is capable of generating images with better perceptual quality. 
\end{itemize}

\section{Related Work}\label{sec: related_work}

\textbf{Traditional ISP.} The traditional Image Signal Processing (ISP) pipeline includes denoising~\cite{dabov2007image, zhang2017beyond}, demosaicing~\cite{gharbi2016deep}, white balancing~\cite{cheng2015beyond}, color correction~\cite{kwok2013simultaneous, rizzi2003new}, gamma correction, and tone mapping~\cite{rana2019deep, liuzhen_01, liuzhen_02}, aiming to convert RAW images to high-quality sRGB images. To date, camera systems typically utilize manual ISP workflows, necessitating experienced engineers to adjust numerous parameters to achieve satisfactory image quality. Sequential execution of multiple subtasks on proprietary hardware can result in accumulated errors. Compared to digital single-lens reflex (DSLR) cameras, mobile devices face hardware limitations that hinder their ability to capture images of comparable quality to those produced by professional DSLRs. 

\textbf{Learning-based ISP.} With the development of deep learning, learning-based approaches~\cite{deepisp, cycleisp, invisp} have intensified to address the hardware constraints and manual adjustments required by traditional ISP methods, which most of these efforts aim to capture the ISP process under controlled conditions where RAW and sRGB training pairs are captured from the same device. To this end, Ignatov \textit{et al.}~\cite{ignatov2020replacing} set a new challenge by tackling the RAW-to-sRGB mapping problem caused by dual-device capture, which involves addressing spatial misalignment and resolution variations. To tackle these challenges, AWNet~\cite{AWNet} explored the potential of wavelet transformation and non-local attention mechanisms in the ISP pipeline. LiteISPNet~\cite{LiteISPNet} created a global color mapping module to address color inconsistency issues and used an aligned loss to compute optical flow between the predicted sRGB images and ground truth. FourierISP~\cite{FourierISP} employed the Fourier prior to separating and refining the color and structural representations. TransformISP~\cite{transformisp} used a color-conditional ISP network with a masked aligned loss to refine color and tone mappings. 
\begin{figure*}[!t]
    \centering
    \includegraphics[width=\linewidth]{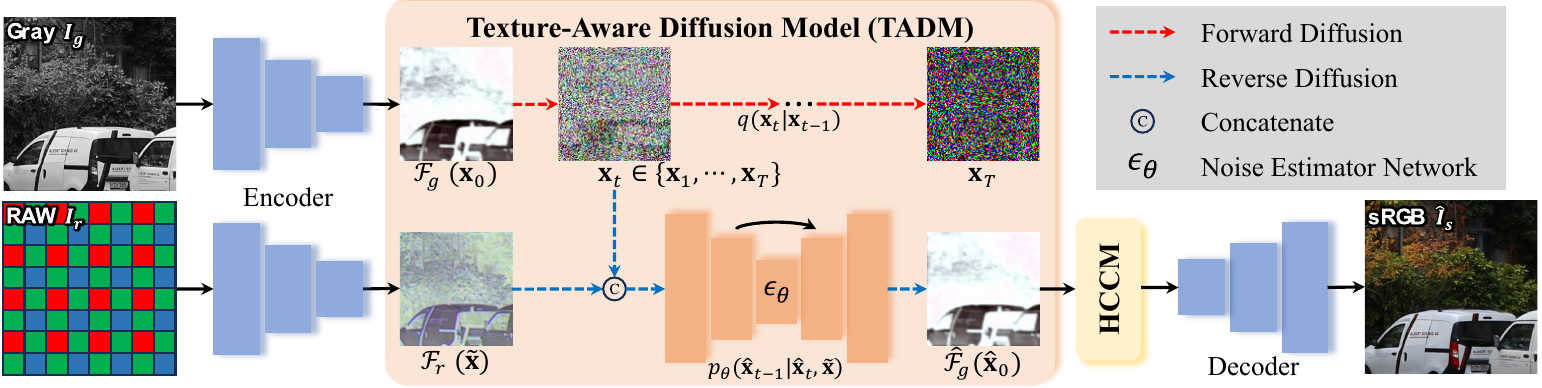}
    \caption{The overall pipeline of our proposed framework. We first employ an encoder $\mathcal{E}(\cdot)$ to convert RAW image $I_{r}$ and grayscale version $I_{g}$ of the sRGB image into latent space denoted as $\mathcal{F}_{r}$ and $\mathcal{F}_{g}$. The encoded feature $\mathcal{F}_{g}$ is taken as the input of the proposed texture-aware diffusion model (TADM) to perform the forward diffusion process. With the guidance of the raw feature $\mathcal{F}_{r}$, we generate the reconstructed gray feature $\hat{\mathcal{F}}_{g}$ from the noised tensor $\mathbf{x}_{t}$ during training, which is replaced by randomly sampled Gaussian noise $\hat{\mathbf{x}}_{T}$ during inference. Finally, we utilize the proposed histogram-guided color consistency module (HCCM) to colorize the generated $\hat{\mathcal{F}}_{g}$ and subsequently send it to a decoder $\mathcal{D}(\cdot)$ to produce the final sRGB result $\hat{I}_{s}$. }
    \label{Pipeline}
\end{figure*}

\textbf{Diffusion Models in Low-Level Vision.}
Diffusion models~\cite{Song_diffusion} are generative models that employ stochastic diffusion processes based on thermodynamics. To date, diffusion models have been widely used in various low-level vision tasks owing to their powerful generative capabilities such as image editing~\cite{image_editing1, image_editing2}, image restoration~\cite{DiffLL, LightenDiffusion, repaint, DDRM}, and image alignment~\cite{FlowDiffuser, DMHomo, RecDiffusion}, while their application to RAW-to-sRGB mappings still needs to be explored. Recently, DiffRAW~\cite{diffraw} introduced a diffusion-based framework for RAW-to-sRGB mappings that utilizes the sRGB images produced by LiteISPNet~\cite{LiteISPNet} as the color-position preserving condition, which would result in the learned color and detail distributions be similar as LiteISPNet. In this work, we proposed a new diffusion-based framework that decouples the RAW-to-sRGB task into grayscale detail reconstruction and color consistency mapping from grayscale to sRGB to achieve more visually satisfactory results. 

\section{Methodology}\label{sec: Methodology}
\subsection{Overview}\label{subsec: overview}
The overall pipeline of our method is illustrated in Fig.~\ref{Pipeline}. Our approach decouples the RAW-to-sRGB mappings into grayscale detail reconstruction and color-consistent mapping from grayscale to sRGB, aiming to transform the RAW images into high-quality sRGB images. Given a RAW image $I_{r} \in \mathbb{R}^{H \times W \times 1}$ and the grayscale version $I_{g} \in \mathbb{R}^{H \times W \times 1}$ of the corresponding sRGB image, we adopt an encoder $\mathcal{E}(\cdot)$, which consists of $k$ cascaded residual blocks where each block downsamples the input by a scale of 2, to transform the input images into latent space denoted as $\mathcal{F}_{r} \in \mathbb{R}^{\frac{H}{2^{k}} \times \frac{W}{2^{k}} \times c}$ and $\mathcal{F}_{g} \in \mathbb{R}^{\frac{H}{2^{k}} \times \frac{W}{2^{k}} \times c}$. Then, we introduce a texture-aware diffusion model (TADM) which leverages the generative ability of diffusion models to transform the RAW feature into the content informative grayscale feature $\mathcal{\hat{F}}_{g}$ with the guidance of the newly proposed texture preservation loss. Subsequently, we propose a histogram-guided color consistency module (HCCM) to colorize the grayscale feature $\mathcal{\hat{F}}_{g}$ into the sRGB feature $\mathcal{\hat{F}}_{s}$, which will be sent to a decoder $\mathcal{D}(\cdot)$ for reconstruction to produce the final sRGB image $\hat{I}_{s}$.

\subsection{Texture-Aware Diffusion Model}\label{subsec: TADM}
RAW-to-sRGB mappings have two critical concerns: local detail recovery and global color mapping. Recently, diffusion models have gained attention for their impressive generative ability, while encountering low-frequency generative bias, such as color and exposure~\cite{DiffLL, diffsuion_bias}. To this end, we present a texture-aware diffusion model (TADM) that focuses on reconstructing the details of sRGB images without attending to the low-frequency color mapping. Our approach follows standard diffusion models~\cite{DDPM, DDIM} that perform forward diffusion and reverse diffusion processes to generate results with fine-grained details. 

\textbf{Forward Diffusion.} We take the gray feature $\mathcal{F}_{g}$ as the initial input $\mathbf{x}_{0}$ for the forward diffusion process, in which a predefined variance schedule $\{\beta_{1},\beta_{2},...,\beta_{T}\}$ is employed to progressively convert $\mathbf{x}_{0}$ into Gaussian noise $\mathbf{x}_{T}\sim{\mathcal{N}(\mathbf{0, I})}$ over $T$ steps, which can be formulated as:
\begin{equation}\label{eq: 1}
    q(\mathbf{x}_{t}|\mathbf{x}_{t-1})=\mathcal{N}(\mathbf{x}_{t};\sqrt{1-\beta_{t}\mathbf{x}_{t-1}}, \beta_{t}\mathbf{I}),
\end{equation}
where $\mathbf{x}_{t}$ indicates the noisy data at time-step $t\in[0,T]$. With parameter renormalization, we can obtain $\mathbf{x}_{t}$ directly from the input $\mathbf{x}_{0}$ and thereby simplify Eq.(\ref{eq: 1}) into a closed expression as $\mathbf{x}_t=\sqrt{\bar{\alpha}_t} \mathbf{x}_0+\sqrt{1-\bar{\alpha}_t} \boldsymbol{\epsilon}_t$, where $\alpha_t = 1 - \beta_t$, $\bar{\alpha}_t=\prod_{i=0}^t \alpha_i$, and $\boldsymbol{\epsilon}_t \sim \mathcal{N}(\mathbf{0},\mathbf{I})$. 

\textbf{Reverse Diffusion.} The reverse diffusion process learns the non-Markovian forward processes that gradually denoise a randomly sampled Gaussian noise $\mathbf{\hat{x}}_{T}\sim\mathcal{N}(\mathbf{0}, \mathbf{I})$ into a sharp result $\mathbf{\hat{x}}_{0}$ conforming to the target data distribution. To strengthen the controllability of the generation procedure, we apply the conditional mechanism~\cite{conditional_dm} to improve the fidelity of the reconstructed results conditioned on the encoded RAW feature $\mathcal{F}_{r}$, denoted as $\Tilde{\mathbf{x}}$. The reverse diffusion process can be formulated as:
\begin{equation}\label{eq: 2}
    p_{\theta}(\hat{\mathbf{x}}_{t-1}|\hat{\mathbf{x}}_{t},\widetilde{\mathbf{x}})=\mathcal{N}(\hat{\mathbf{x}}_{t-1};\boldsymbol{\mu}_{\theta}(\hat{\mathbf{x}}_{t},\widetilde{\mathbf{x}},t),\sigma_{t}^{2}\mathbf{I}),
\end{equation}
where $\sigma_{t}^{2}=\frac{1-\overline{\alpha}_{t-1}}{1-\overline{\alpha}_{t}}\beta_{t}$ is the variance and $\boldsymbol{\mu}_{\theta}(\hat{\mathbf{x}}_{t},\widetilde{\mathbf{x}},t)=\frac{1}{\sqrt{\alpha_{t}}}(\hat{\mathbf{x}}_{t}-\frac{\beta_{t}}{\sqrt{1-\overline{\alpha}_{t}}}\boldsymbol{\epsilon}_{\theta}(\hat{\mathbf{x}}_{t},\widetilde{\mathbf{x}},t))$ is the mean value. 

In the training phase, instead of optimizing the parameters $\theta$ of the network $\boldsymbol{\epsilon}_{\theta}$ to promote the estimated noise vector close to Gaussian noise, we follow~\cite{FlowDiffuser, DMHomo} to generate the sharp gray feature $\mathbf{\hat{x}}_{0}$, i.e., $\hat{\mathcal{F}}_{g}$ and employ the content loss $\mathcal{L}_{con}$ for optimization as:
\begin{equation}\label{eq: 3}
    \mathcal{L}_{con}=||\hat{\mathbf{x}}_{0}-\mathbf{x}_{0}||_{2},
\end{equation}
where $\hat{\mathbf{x}}_{0}$ is estimated from the disturbed noise data as:
\begin{equation}\label{eq: 4}
   \hat{\mathbf{x}}_{0}=\frac{1}{\sqrt{\bar{\alpha}_t}}\left(\mathbf{x}_t-\sqrt{1-\bar{\alpha}_t} \boldsymbol{\epsilon}_\theta\left(\mathbf{x}_t, \tilde{\mathbf{x}}, t\right)\right).
\end{equation}

Furthermore, we introduce a texture enrichment loss $\mathcal{L}_{tel}$ to enable the reconstructed features to contain detailed texture information similar to the original input. Specifically, we employ the traditional Canny edge detector to extract the corresponding texture maps of the generated feature $\hat{\mathcal{F}}_{g}$ and the original gray feature $\mathcal{F}_{g}$ before non-maximum suppression, denoted as $\hat{\mathbf{T}}_{g} = \operatorname{Canny}(\hat{\mathcal{F}}_{g})$ and $\mathbf{T}_{g} = \operatorname{Canny}(\mathcal{F}_{g})$. Thus, the $\mathcal{L}_{tel}$ aims to constrain the texture similarity to prompt the diffusion model to generate $\hat{\mathcal{F}}_{g}$ with more intricate texture details as:
\begin{equation}\label{eq: 5}
    \mathcal{L}_{tel}=||\hat{\mathbf{T}}_{g}-\mathbf{T}_{g}||_{1}.
\end{equation}

Overall, the object function used to optimize our TADM is rewritten as $\mathcal{L}_{diff}=\mathcal{L}_{con}+\lambda_1 \mathcal{L}_{tel}$. During inference, we derive the restored feature $\hat{\mathcal{F}}_{g}$ from the learned distribution through the reverse diffusion process with the implicit sampling strategy~\cite{DDIM}. 

\subsection{Histogram-Guided Color Consistency Model}\label{subsec: HCCM}
\begin{figure}[!t]
    \centering
    \includegraphics[width=\linewidth]{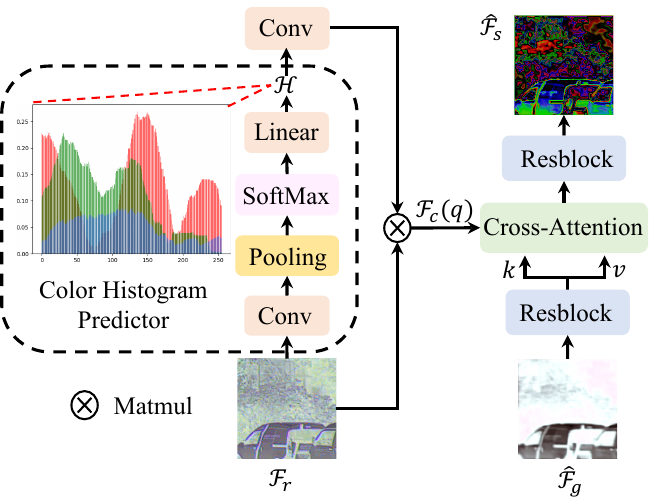}
    \caption{The detailed architecture of our proposed histogram-guided color consistency module. }
    \label{HCCM}
\end{figure}

The Bayer filter array (BFA) captures color information by processing specific color arrangements within individual channels, which makes the RAW to sRGB transformation suffer from color disparities and unstable color mapping. To this end, with our decoupled framework, we introduce a histogram-guided color consistency module (HCCM) that utilizes the color histogram~\cite{color_histogram} as guidance to transform the gray feature generated by our TADM to sRGB feature with vivid color, as illustrated in Fig.~\ref{HCCM}. 

Specifically, the RAW feature $\mathcal{F}_{r}$ is taken as input to our designed color histogram predictor $\mathcal{CHP}(\cdot)$ to predict the color histogram $\mathcal{H}\in\mathcal{R}^{N\times256}$ of sRGB distribution, where $N=3$ indicates the number of color channels in sRGB space (i.e., R, G, and B) and 256 aligns with the range of pixel values. However, since color histogram primarily describes the proportion of different colors across the entire image without accounting for spatial arrangement. Therefore, we utilize the RAW feature along with $\mathcal{H}$ to extract the position-specific color feature $\mathcal{F}_{c}$ as:
\begin{equation}
    \mathcal{F}_{c}=\operatorname{Conv}(\mathcal{H}) \times \mathcal{F}_{r}^{'}, \mathcal{H} = \mathcal{CHP}(\mathcal{F}_{r}),
\end{equation}
where $\mathcal{F}_{r}^{'}$ is the reshaped RAW feature to satisfy dimension alignment and $\times$ denotes the matrix multiplication. Subsequently, we adopt a cross-attention layer to leverage the estimated position-specific color feature $\mathcal{F}_{c}$ to colorize the grayscale feature into sRGB feature $\hat{\mathcal{F}}_{s}$ enriched with detailed information and consistent color, in which the $\mathcal{F}_{c}$ is taken as query vector $q$ while key $k$ and value $v$ vectors are calculated from original grayscale feature $\hat{\mathcal{F}}_{g}$. 

To facilitate the $\mathcal{CHP(\cdot)}$ to predict more accurate color histograms for grayscale to sRGB transformation, we design a color consistency loss $\mathcal{L}_{ccl}$ to optimize the estimated $\mathcal{H}$ align with the color histogram $\mathcal{H}_{s}$ of encoded ground-truth sRGB feature $\mathcal{F}_{s}$, which is formulated as:
\begin{equation}
    \mathcal{L}_{ccl}=||\mathcal{H}-\mathcal{H}_{s}||_{2}.
\end{equation}
Moreover, a feature loss $\mathcal{L}_{fea}$ is also adopted to constrain the reconstructed sRGB feature as:
\begin{equation}
\mathcal{L}_{fea} = ||\hat{\mathcal{F}}_{s}-\mathcal{F}_{s}||_{2}.
\end{equation}
Overall, the objective function used to optimize the HCCM is formulated as $\mathcal{L}_{hccm} = \mathcal{L}_{fea} + \lambda_2 \mathcal{L}_{ccl}$.

\subsection{Network Training}\label{subsec: network_training}
Our approach employs a two-stage training strategy. In the first stage, we use paired RAW-sRGB images to train the encoder $\mathcal{E}(\cdot)$ and decoder $\mathcal{D}(\cdot)$, while freezing the parameters of the diffusion model and HCCM. The encoder and decoder are optimized with the content loss $\mathcal{L}_{stage1}$ as:
\begin{equation}
    \mathcal{L}_{stage1} = ||I - \mathcal{D}(\mathcal{E}(I))||_2,
\end{equation}
where $I$ denotes the input RAW, sRGB, and grayscale images. In the second stage, we optimize the TADM and HCCM simultaneously through $\mathcal{L}_{stage2} = \mathcal{L}_{diff} + \mathcal{L}_{hccm}$ while freezing the parameters of the encoder and decoder. 
\begin{table*}[!t]
    \centering
    \resizebox{\linewidth}{!}{
    \begin{tabular}{l|c|ccc|ccc|ccc}
    \toprule
    \multirow{2}[2]{*}{Method} & \multirow{2}[2]{*}{Time (ms)} & \multicolumn{3}{c|}{ZRR (Original GT)} & \multicolumn{3}{c|}{ZRR (Align GT with RAW)} & \multicolumn{3}{c}{MAI} \\
          & & PSNR $\uparrow$ & SSIM $\uparrow$ & LPIPS $\downarrow$ & PSNR $\uparrow$ & SSIM $\uparrow$ & LPIPS $\downarrow$ & PSNR $\uparrow$ & SSIM $\uparrow$ & LPIPS $\downarrow$ \\
    \midrule
    PyNet & 62.7 & 21.19 & 0.747 & 0.193 & 22.73 & 0.845 & 0.152 & 23.81 & 0.848 & 0.139 \\
    AWNet-R & 55.7 & 21.42 & 0.748 & 0.198 & 23.27 & 0.854 & 0.151 & 24.53 & 0.872 & 0.136 \\
    AWNet-D & 62.7 & 21.53 & 0.749 & 0.212 & 23.38 & 0.850 & 0.164 & 24.64 & 0.866 & 0.147 \\
    MW-ISPNet & 110.5 & 21.42 & \underline{0.754} & 0.213 & 23.07 & 0.848 & 0.165 & 25.02 & 0.885 & 0.133 \\
    LiteISPNet & 23.3 & 21.55 & 0.749 & 0.187 & 23.76 & 0.873 & 0.133 & 24.90 & 0.877 & 0.123 \\
    FourierISP & 25.0  & \underline{21.65} & \textbf{0.755} & 0.182 & \underline{23.93} & \underline{0.874} & \underline{0.124} & \underline{25.37} & \underline{0.891} & \underline{0.072} \\
    DiffRAW  & -  & 21.31 & 0.743 & \textbf{0.145} & -  & - & - & - & - & - \\
    ISPDiffuser (Ours) & 490.0 &\textbf{21.77} & \underline{0.754} & \underline{0.157} & \textbf{24.09} & \textbf{0.881} & \textbf{0.111} & \textbf{25.64} & \textbf{0.894} & \textbf{0.071} \\
    \bottomrule
    \end{tabular}}
    \caption{Quantitative comparisons on the ZRR~\cite{PyNet} and MAI~\cite{MAI} test sets. The best results are highlighted in \textbf{bold} and the second best results are in \underline{underlined}. `-' indicates the results are unavailable since the source code of DiffRAW~\cite{diffraw} has not been publicly released. `Time' denotes the average time cost when performing inference on the ZRR test set, where the images are with 448$\times$448 resolution. }
    \label{tab: Quantitative Comparison}
\end{table*}

\begin{table}[!t]
    \centering
     \resizebox{\linewidth}{!}{
    \begin{tabular}{l|cc|cc}
    \toprule
    \multirow{2}[2]{*}{Method} & \multicolumn{2}{c|}{ZRR} & \multicolumn{2}{c}{MAI} \\
    & MUS. $\uparrow$ & TOP. $\uparrow$ & MUS. $\uparrow$ & TOP. $\uparrow$ \\
    \midrule
    PyNet & 43.796 & 0.362 & 39.823 & 0.445 \\
    AWNet-R & 43.441 & 0.355 & 40.211 & 0.441 \\
    AWNet-D  & 45.100 & 0.362 & 39.839 & 0.432 \\
    MW-ISPNet & 42.448 & 0.340 & 40.652 & \underline{0.449} \\
    LiteISPNet & \underline{47.310} & \underline{0.370} & 40.365 & 0.445 \\
    FourierISP & 44.534 & 0.369 & \underline{47.614} & \textbf{0.535} \\
    ISPDiffuser (Ours)  & \textbf{50.117} & \textbf{0.392} & \textbf{48.032} &  \textbf{0.535} \\
    \bottomrule
    \end{tabular}
    }
    \caption{Non-reference perceptual metric comparisons on the ZRR~\cite{PyNet} and MAI~\cite{MAI} test sets. The best results are highlighted in \textbf{bold} and the second best results are in \underline{underlined}. `MUS.'and `TOP.' denote MUSIQ and TOPIQ. }
    \label{tab: IQA_comparision}
\end{table}

\section{Experiments}\label{sec: experiment}
\subsection{Experimental Settings}\label{subsec: experimental_settings}
\textbf{Implementation Details.} We implement the proposed method with PyTorch on four NVIDIA RTX 2080Ti GPUs, where the batch size and patch size are set to $16$ and $256\times256$. We employ the Adam optimizer~\cite{Adam} for optimization with the initial learning rate set to $1\times10^{-4}$ and decay by a factor of 0.8 in both two stages. The feature downsampling scale $k$ is set to 2. The hyperparameters $\lambda_1$ and $\lambda_2$ are both set to 0.01. For our TADM, the U-Net~\cite{UNet} architecture is adopted as the noise estimator network with the time step $T$ and sampling step $S$ set to 1000 and 25 for the forward diffusion and reverse diffusion process, respectively. 

\textbf{Datasets.} We have conducted experiments on two publicly available benchmarks including ZRR~\cite{PyNet} and MAI~\cite{MAI} datasets. The ZRR dataset contains 46.8k RAW-sRGB image pairs for training and 1.2k pairs for evaluation, where the RAW images are captured by Huawei P20 and the sRGB images are captured by Canon camera. Meanwhile, the MAI dataset concentrates on mapping the RAW images captured by Sony IMX586 to sRGB distributions of the Fuji camera. Since the test set of the MAI dataset lacks GT sRGB images, we follow~\cite{FourierISP} to split the training set into 90\% for training (21.7k pairs) and 10\% for evaluation (2.4k pairs). Notably, the RAW images in the ZRR dataset are 10-bit, while those in the MAI dataset are 12-bit.

\textbf{Metrics.} We adopt two full-reference distortion metrics PSNR and SSIM~\cite{SSIM}, and a perceptual metric LPIPS~\cite{LPIPS} for evaluation. Moreover, two non-reference perceptual metrics MUSIQ~\cite{MUSIQ} and TOPIQ~\cite{topiq} are adopted to measure the visual quality of generated images. 

\subsection{Comparison with Existing Methods}\label{subsec: comparisons}
\textbf{Comparison Methods.}
In this section, we compare the proposed method with existing state-of-the-art methods, including PyNet~\cite{PyNet}, AWNet~\cite{AWNet}, MW-ISPNet~\cite{MW_ISPNet}, LiteISPNet~\cite{LiteISPNet}, FourierISP~\cite{FourierISP}, and DiffRAW~\cite{diffraw}. Note that we follow~\cite{LiteISPNet} which uses two models of AWNet, i.e., AWNet (RAW) denoted as AWNet-R, and AWNet (demosaic) denoted as AWNet-D, for comparison. Moreover, the metrics of DiffRAW are adopted from its associated publication since the source code is unavailable. 

\textbf{Quantitative Comparison.}
We compare our method with all comparison methods on the ZRR~\cite{PyNet} and MAI~\cite{MAI} test sets. Since the image pairs in the ZRR dataset present spatial misalignment, we follow~\cite{LiteISPNet} that calculate the metrics using the original ground truth sRGB images denoted as ``Original GT" and adopt the optical flow estimation method PWCNet~\cite{Pwcnet} to align the image pairs for evaluation as ``Align GT with RAW". 
\begin{figure*}[!t]
    \centering
    \includegraphics[width=\linewidth]{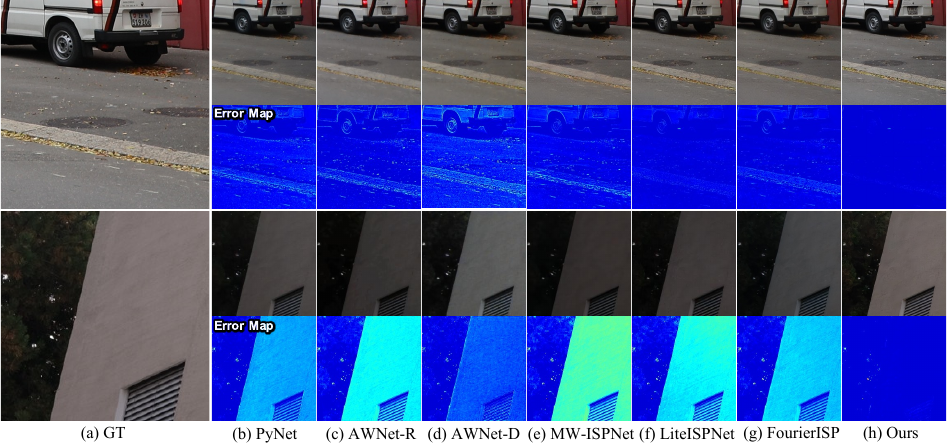}
    \caption{Qualitative comparison of our method and competitive methods on the ZRR dataset~\cite{PyNet} (row 1) and MAI dataset~\cite{MAI} (row 2). The error maps represent the content difference between the generated sRGB images and the GT images, the darker the better. Best viewed by zooming in. }
    \label{fig: comparision}
\end{figure*}

As reported in Table~\ref{tab: Quantitative Comparison}, our method outperforms the second-best method FourierISP by 0.12dB and 0.16dB in terms of PSNR under both evaluation modalities of the ZRR dataset. For SSIM and LPIPS, our method is slightly inferior to FourierISP and DiffRAW under the setting of ``Original GT" on the ZRR dataset due to the misalignment between original RAW-sRGB image pairs, while achieving the best results with 0.007 and 0.013 improvements in another mode after alignment, which indicates our method can produce images with satisfactory visual quality and is more appropriate for RAW-to-sRGB mappings. For the MAI dataset, our method obtains state-of-the-art performance in all three metrics with 0.27dB improvement in terms of PSNR, 0.003 improvement in SSIM, and 0.001 improvement in LPIPS, respectively, which proves the strong generalization ability of our method. To further validate the effectiveness of our method, we adopt two non-reference perceptual metrics to measure the visual quality of our generated sRGB images on the ZRR and MAI test sets. As illustrated in Table~\ref{tab: IQA_comparision}, our method achieves the best performance in both two metrics on the ZRR and MAI test sets with a remarkable improvement in MUSIQ, which proves that our method is capable of generating images with better perceptual quality. 

\textbf{Qualitative Comparison.} We present qualitative comparisons of our method and competitive methods on the ZRR and MAI test sets, as illustrated in Fig.~\ref{fig: comparision}. For better visualization, we provide the error maps beneath each corresponding image that indicate the content discrepancies between the generated sRGB images and GT images. As we can see previous methods, such as PyNet and both variants of AWNet, struggle to preserve color accuracy and detail sharpness. MW-ISPNet, LiteISPNet, and FourierISP exhibit enhanced performance, yet they continue to encounter limitations in preserving texture richness and ensuring consistent color reproduction. These shortcomings become especially apparent in areas with complex illumination or intricate details, where these methods often produce unexpected artifacts or lose essential information. In contrast, our method can generate results that exhibit high visual fidelity to the ground truth. By effectively capturing and reproducing fine textures, such as intricate patterns and subtle gradients, while maintaining accurate and vibrant color representation, our approach achieves more lifelike, detailed, and visually compelling reconstructions. The superiority of our method is further substantiated by error maps, which reveal minimal discrepancies between our outputs and GT images, underscoring the effectiveness of our approach. 

\subsection{User Study}\label{subsec: user_study}
\begin{figure}[!t]
    \centering
     \includegraphics[width=\linewidth]{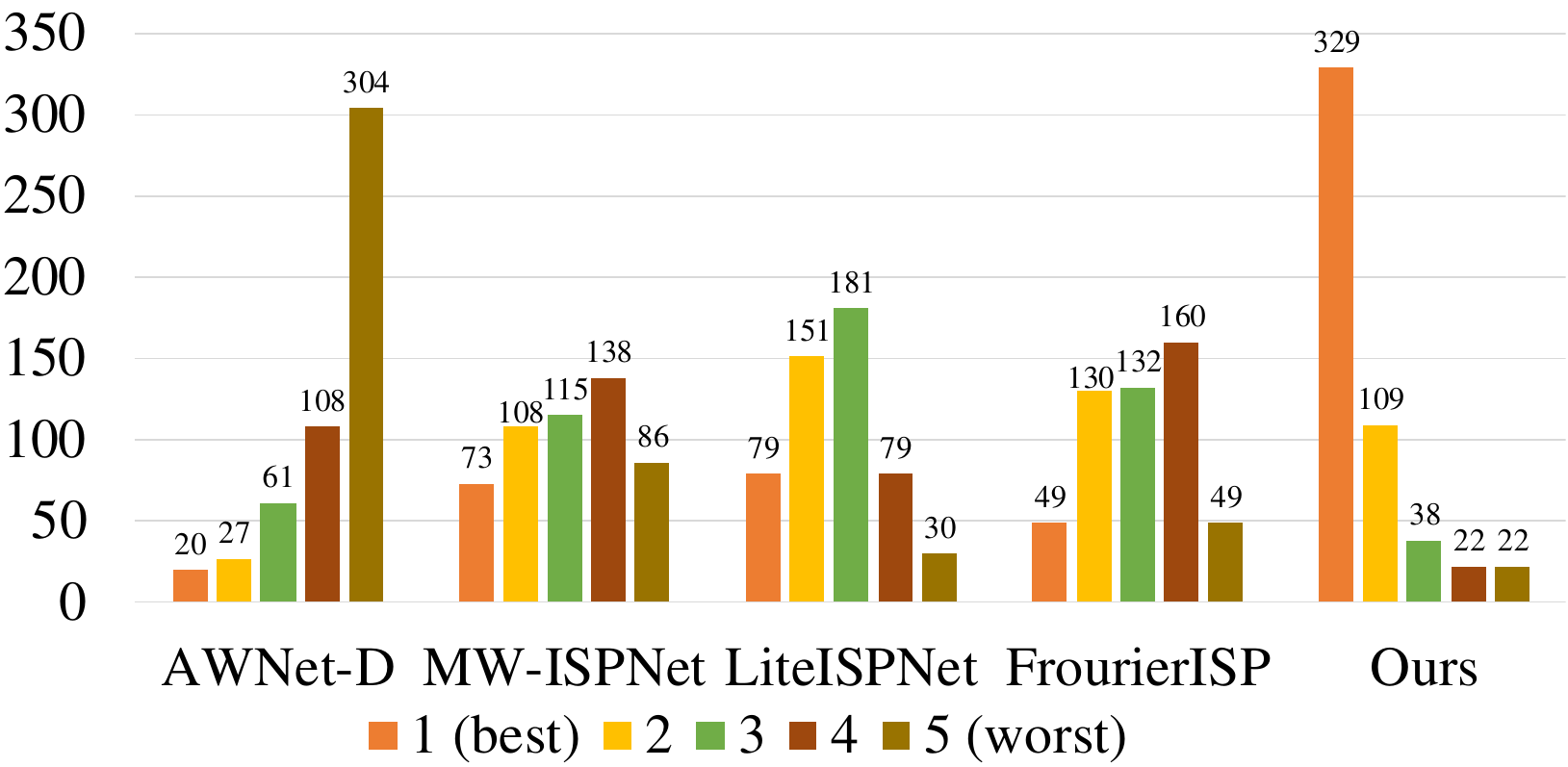}
    \caption{Score distributions of user study, where the ordinate axis records the rating frequency received from the 26 participants. Our method receives more ``best” ratings.}
    \label{fig: user_study}
\end{figure}

We further conduct a user study to compare the proposed method with four competitive methods including AWNet-D, MW-ISPNet, LiteISPNet, and FourierISP. We randomly select 20 images from the ZRR and MAI test sets and invited 26 participants to measure the subjective preference of the above methods. For each case, the input RAW data and the generated sRGB results of the five methods are shown at the same time. The participants are required to rank the results from 1 (best) to 5 (worst) based on the perceptual quality including local details and global color. Fig.~\ref{fig: user_study} illustrates the rating distributions of different methods, where our method receives more `best' ratings, proving our results are more preferred by human subjects. 

\subsection{Ablation Study}\label{subsec: ablations}
In this section, we conduct a series of ablation studies to validate the effectiveness of newly proposed components in our method. The quantitative results on the ZRR~\cite{PyNet} test set under the setting of ``Original GT" are reported in Table~\ref{tab: ablation framework&HCCM} and Table~\ref{tab: ablation loss}. 

\subsubsection{Our Framework.}
To validate the effectiveness of our decoupled framework, we employ the TADM only to form a baseline that takes the sRGB image as input and the RAW image as the condition to directly perform RAW-to-sRGB mappings in the image space, i.e., $k=0$. We can observe that the baseline without decoupled suffers from poor sharpness with incorrect exposure and color distortion as shown in Fig.~\ref{fig:ablation_1}(b), caused by the simultaneous handling of detail reconstruction and color mapping. In contrast, our decoupled framework that separates grayscale texture reconstruction from colorization delivers sharper details and presents vivid color as shown in Fig.~\ref{fig:ablation_1}(d), presenting overall performance improvements as reported in rows 1 and 4 of Table~\ref{tab: ablation framework&HCCM}. 

\subsubsection{HCCM Module.} 
\begin{figure}[!t]
    \centering
    \includegraphics[width=\linewidth]{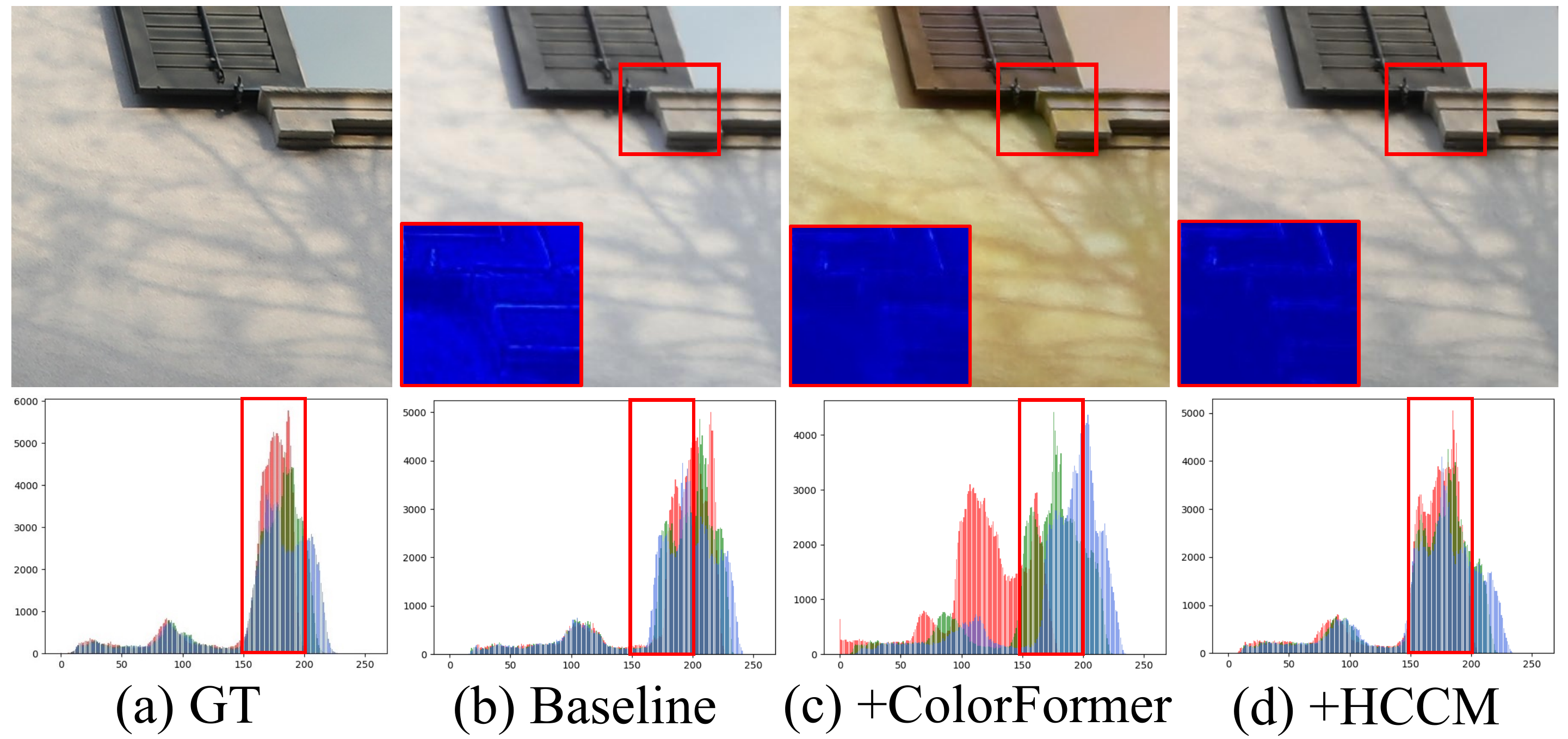}
    \caption{Visual results of the ablation study about our proposed framework and HCCM module. The second row showcases the color histogram of the image. }
    \label{fig:ablation_1}
\end{figure}

To validate the effectiveness of our proposed histogram-guided color consistency module, we replace it with two established automatic image colorization methods ColorFormer~\cite{colorformer} and DDColor~\cite{ddcolor}, which achieve state-of-the-art performance in transforming grayscale images to sRGB images. As reported in rows 2-4 of Table~\ref{tab: ablation framework&HCCM}, our method with the HCCM module outperforms these colorization methods in terms of all metrics. As illustrated in Fig.~\ref{fig:ablation_1}(c) and (d), the sRGB image generated by ColorFormer encounters overall color discrepancies, whereas our HCCM is capable of generating results with more accurate and consistent color. 
\begin{table}[!t]
    \centering
     \resizebox{\linewidth}{!}{
    \begin{tabular}{l|c|ccc}
    \toprule
         Methods & Decouple & PSNR $\uparrow$ &SSIM $\uparrow$ & LPIPS $\downarrow$ \\
         \midrule
         Baseline & & 20.12 & 0.731 & 0.208\\
         +DDColor & \checkmark & 18.83 & 0.711 & 0.293\\
         +ColorFormer & \checkmark & 19.24 & 0.716 & 0.278\\
         +HCCM & \checkmark &20.93 & 0.740 & 0.204\\
         \bottomrule
    \end{tabular}
    }
    \caption{Ablation studies of our proposed decoupled framework and histogram-guided color consistency module. }
    \label{tab: ablation framework&HCCM}
\end{table}

\subsubsection{Loss Functions.} 
\begin{figure}[!t]
    \centering
     \includegraphics[width=\linewidth]{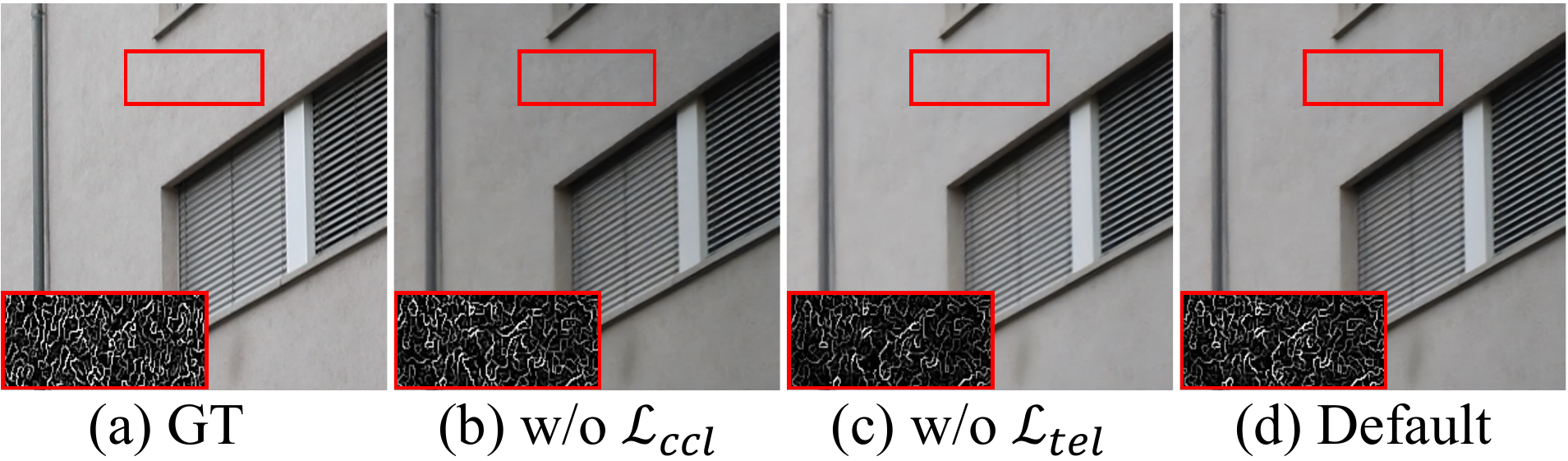}
    \caption{Visual results of the ablation study about our proposed texture enrichment loss $\mathcal{L}_{tel}$ and color consistency loss $\mathcal{L}_{ccl}$. `w/o' denotes without. }
    \label{fig:ablation_2}
\end{figure}

To validate the effectiveness of our proposed texture enrichment loss $\mathcal{L}_{tel}$ and color consistency loss $\mathcal{L}_{ccl}$, we conduct experiments by individually removing each component from the default setting, where the quantitative results are reported in Table~\ref{tab: ablation loss}. As shown in row 1, the removal of the above two objective functions results in overall performance degradation.  The $\mathcal{L}_{ccl}$ is designed to facilitate the HCCM transforming the grayscale feature into the sRGB feature with vivid color, which is beneficial in enhancing the visual fidelity of the restored images, thereby leading to the overall performance improvement. As illustrated in Fig.~\ref{fig:ablation_2}(b) and (d), the inclusion of $\mathcal{L}_{ccl}$ for optimization is capable of correcting global color distortion, leading to more satisfactory results. The incorporation of $\mathcal{L}_{tel}$ is helpful to generate results with shaper details thus achieving noticeable improvements in terms of the distortion metrics. As illustrated in Fig.~\ref{fig:ablation_2}(c) and (d), the $\mathcal{L}_{tel}$ is helpful in generating images with enriched texture information. 

\begin{table}[!t]
    \centering
     \resizebox{0.75\linewidth}{!}{
    \begin{tabular}{cc|ccc}
    \toprule
        $\mathcal{L}_{tel}$&$\mathcal{L}_{ccl}$& PSNR $\uparrow$ & SSIM $\uparrow$ & LPIPS $\downarrow$ \\
         \midrule
                    &            & 21.30 & 0.736 & 0.161\\
         \checkmark &            & 21.33 & 0.751 & 0.156\\
                    & \checkmark & 21.62 & 0.750 & 0.158\\
         \checkmark & \checkmark & 21.77 & 0.754 & 0.157 \\
         \bottomrule
    \end{tabular}}
    \caption{Ablation studies of our proposed texture enrichment loss $\mathcal{L}_{tel}$ and color consistency loss $\mathcal{L}_{ccl}$. }
    \label{tab: ablation loss}
\end{table}

\subsection{Litimations}\label{sec: limitations}
Although our method effectively transforms RAW images into DSLR-quality sRGB images, the generalizability to diverse weather, lighting, and devices is limited by the training data from specific cameras. Moreover, since diffusion-based methods rely on iterative denoising of Gaussian noise, our method shows inferior efficiency compared to some lightweight methods which can be applied to the camera to replace the traditional ISP pipeline, as reported in Table~\ref{tab: Quantitative Comparison}. In the future, we will explore more effective sampling strategies, such as DPM-Solver~\cite{Dpm-solver} and consistency model~\cite{consistency_model}, to improve inference efficiency and investigate the effectiveness of our method.

\section{Conclusion}\label{sec: conclusion}
We have presented ISPDiffuser, a diffusion-based decoupled framework that separates the RAW-to-sRGB mapping into detail reconstruction in grayscale space and color consistency mapping from grayscale to sRGB. Technically, we propose a texture-aware diffusion model that leverages the generative ability of diffusion models to perform grayscale detail reconstruction without concern for color information, where a texture enrichment loss is further proposed to promote the diffusion model to generate more intricate texture details. Subsequently, we construct a histogram-guided color consistency module that utilizes the traditional color histogram as guidance to learn accurate color information for grayscale to sRGB color consistency mapping, with a color consistency loss designed to constrain the learned color information being close to standard DSLR color distribution.  Extensive experiments demonstrate that the proposed ISPDiffuser outperforms existing state-of-the-art competitors both quantitatively and qualitatively. 
\section{Acknowledgments}
This work was supported in part by the National Natural Science Foundation of China (NSFC) under Grant Nos. 62271334, 62372091, 62071097, and in part by the Sichuan Science and Technology Program under Grant Nos. 2023NSFSC0462, 2023NSFSC0458, 2023NSFSC1972.
\bibliography{aaai25}

\end{document}